# Predicting Traffic Congestion at Urban Intersections Using Data-Driven Modeling

Tara Kelly, Jessica Gupta

Tarakelly963@gmail.com, jessicagg33@gmail.com

**Abstract:** Traffic congestion at intersections is a significant issue in urban areas, leading to increased commute times, safety hazards, and operational inefficiencies. This study aims to develop a predictive model for congestion at intersections in major U.S. cities, utilizing a dataset of trip-logging metrics from commercial vehicles across 4,800 intersections. The dataset encompasses 27 features, including intersection coordinates, street names, time of day, and traffic metrics (Kashyap et al., 2019). Additional features, such as rainfall/snowfall percentage, distance from downtown and outskirts, and road types, were incorporated to enhance the model's predictive power. The methodology involves data exploration, feature transformation, and handling missing values through low-rank models and label encoding. The proposed model has the potential to assist city planners and governments in anticipating traffic hot spots, optimizing operations, and identifying infrastructure challenges.

**Keywords:** Machine Learning, Big Data, Data Mining, Data Modeling, Data Security

## 1. Introduction

Urban traffic congestion is a significant challenge faced by modern cities, impacting commute times, safety, and overall quality of life for residents (Weisbrod et al., 2003). Intersections, which serve as crucial points of convergence for vehicular traffic, are particularly susceptible to congestion, leading to stop-and-go patterns and increased travel times (Gazis, 2002). Predicting congestion at intersections can provide valuable insights for city planners and governments, enabling them to implement strategies for optimizing traffic flow, enhancing infrastructure, and improving the overall transportation system (Wang et al.,

2016).

This study aims to develop a predictive model for congestion at intersections in major U.S. cities, leveraging a comprehensive dataset of trip-logging metrics from commercial vehicles. The dataset encompasses a wide range of features, including intersection coordinates, street names, time of day, and traffic metrics, providing a rich source of information for identifying potential congestion patterns.

## 2. Literature Review

Numerous studies have explored the application of machine learning and predictive modeling techniques to address urban traffic challenges. Researchers have employed various approaches, including neural networks (Huang et al., 2014), decision trees, and ensemble methods, to predict traffic flow and congestion patterns. However, many of these studies have focused on specific cities or regions, limiting their generalizability to diverse urban environments. Additionally, the incorporation of contextual features, such as weather conditions, road types, and proximity to central business districts, has been shown to enhance the predictive power of traffic models. By accounting for these external factors, models can better capture the complexities of urban traffic dynamics and provide more accurate predictions.

Neural networks, particularly deep learning architectures, have gained significant attention in traffic prediction due to their ability to capture non-linear relationships and extract relevant features from raw data (Lv et al., 2015). Traffic Flow Prediction with Big Data: A Deep Learning Approach by Lv et al. proposed a stacked autoencoder model for traffic flow prediction, demonstrating improved performance compared to traditional methods. Decision trees and ensemble methods, such as random forests and gradient boosting, have also been widely employed in traffic prediction tasks (Vlahogianni et al., 2014). These approaches are capable of handling non-linear relationships and can provide interpretable models, making them valuable for understanding the underlying factors influencing traffic patterns.

While these studies have contributed to the advancement of traffic prediction models, there is a need for more comprehensive research that considers diverse urban environments and integrates multiple data sources to capture the full complexity of traffic dynamics. Additionally, the development of robust and scalable models that can be effectively deployed in real-world intelligent transportation systems remains an ongoing challenge. Combined with computer vision for human poses (Zhu et al., 2021), traffic predicting models could potentially benefit autonomous driving technology.

# 3. Methodology

Data Exploration and Feature Engineering: The study commenced with a comprehensive exploration of the dataset provided on Kaggle (Kashyap et al., 2019), examining the distributions, patterns, and relationships among the various features. This analysis informed the feature engineering process, where additional features, such as the percentage of rainfall/snowfall, distance from downtown and outskirts, and road types, were derived and incorporated into the dataset.

Feature Transformation and Handling Missing Values: To ensure the integrity of the data and improve model performance, appropriate techniques were employed for feature transformation and handling missing values. Continuous features were scaled using min-max normalization, while categorical features underwent one-hot encoding (Kuhn & Johnson, 2013). For missing values in the street name features, low-rank matrix factorization models were utilized, leveraging the characteristics of the data to impute missing entries (Hastie et al., 2009).

Model Development and Evaluation: A variety of machine learning algorithms were explored for developing the predictive model, including linear regression, decision trees, random forests, and neural networks. The models were trained on a subset of the data (70%) and evaluated using appropriate metrics, such as mean squared error (MSE) for regression tasks, and accuracy for classification tasks (Hastie et al., 2009).

K-fold cross-validation (k=5) was employed to ensure the robustness and generalizability of the models. Hyperparameter tuning was performed using randomized search to optimize the model's performance further (Bergstra & Bengio, 2012).

# 4. Results and Analysis

## 5.1 Dataset Description and Additional Features

The dataset we use contains trip-logging metrics by commercial vehicles from about 4,800 unique intersections in four major cities. It contains 27 features and more than 857k entries. These features include city, intersection ID's, coordinates, entry and exit street names, hour of day, weekend or not, month, direction of entries and exits and percentiles for total time stopped, distance from first stop and time from first stop. Most of the time records are zeros. We have much more intersection data for city 2 and 3, less for city 0 and 1. The dataset contains missing values only for the two features containing street names.

Apart from existing features, we also conducted further research and added additional features which might help in predicting congestion. The first feature we included was the percentage of rainfall/snowfall during that time. Another feature we included was the distance of that intersection to that city's downtown area.

Furthermore, we also included the road type for a particular intersection. This road type includes information on whether the road is a street, lane, boulevard, broad, drive etc. Additionally, we included a feature that contains the distance of the particular intersection from the outskirts.

## 5.2 Data Exploration, Feature Transformation and Handling Missingness

We analyze how many roads are linked to a particular intersection. Some roads might be one way and so the number of entry and exits are counted separately. All intersections have at least one entry and at least one exit. The number of entry and exit streets along with the difference in exit and entry streets for intersections are added to our training set.

For imputing missing values, we first split data based on cities and then use Low Rank Models. The street names are label encoded and then imputed using different loss and regularization functions. Alternatively, we also tried using multinomial and one vs. all loss for nominal variables. However, these techniques did not give errors within a reasonable margin and hence the missing names were encoded as ‘Unknown’.

Since the directions are related to each other, they are encoded to numerical values. Although there is a city name feature, we check if there are roads in between the cities etc. or mislabeled city name in the data. By using K-means method on the longitude and latitude, we easily cluster the data into four groups, and the graph shows that they are perfectly clustered (Figure 1). This means that we do not have to deal with issues like roads between cities etc.

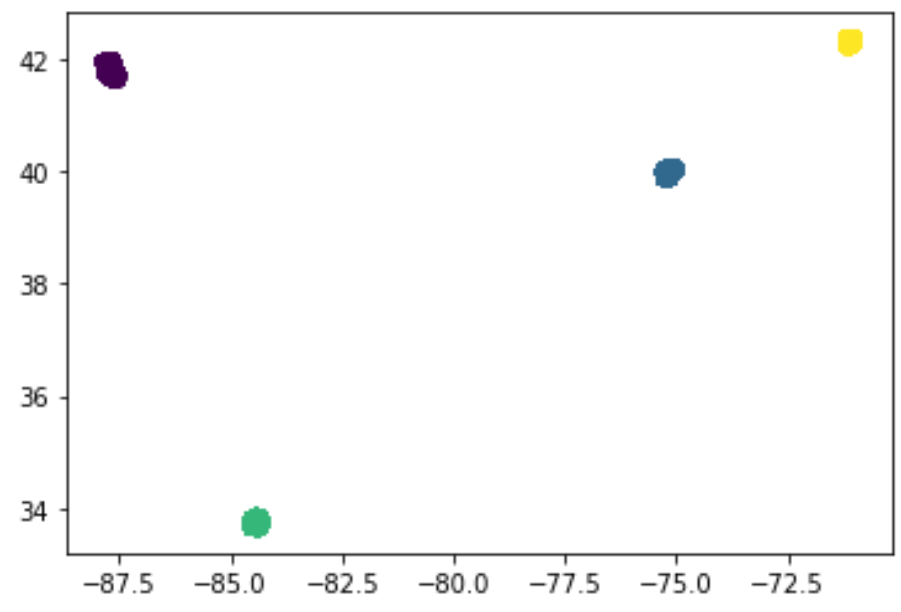

*Figure 1: Four data clusters by using K-means method.*

It is further observed that even with the separation of time, the data is highly unbalanced (Figure 2). Thus, we try to find the busy streets first. We restrict it to be at least 30 minutes waiting time on average.

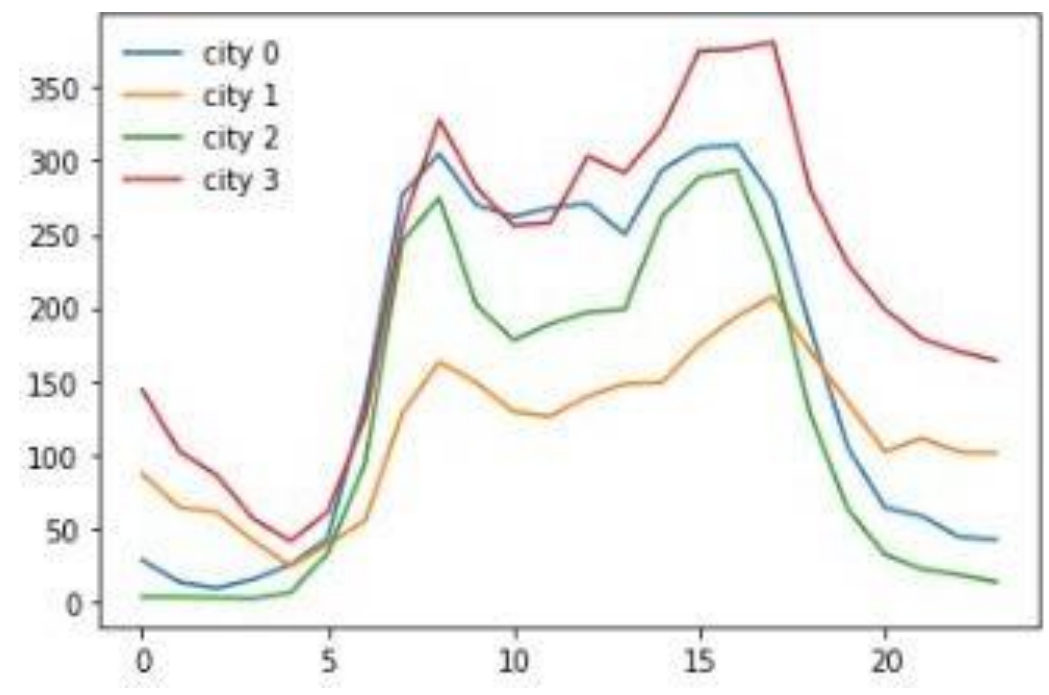

*Figure 2: Number of busy intersections across 24 hours span.*

Some interesting observations from the above data and chart:

- Even though for city 0 we only have 973 unique intersections in our data, we see in its peak time, about a third of its intersections are busy. For city 1 we only have 377 unique intersections and at its peak time about half of them are busy
- The number of busy streets is closer to each other despite the number of total intersections in that city. A more stringent definition of 'busy' may bring these curves closer.

Visualizing the busy intersections in for the cities of Atlanta (Figure 3), Boston (*Figure 4*), Chicago (Figure 5), and Philadelphia (Figure 6):

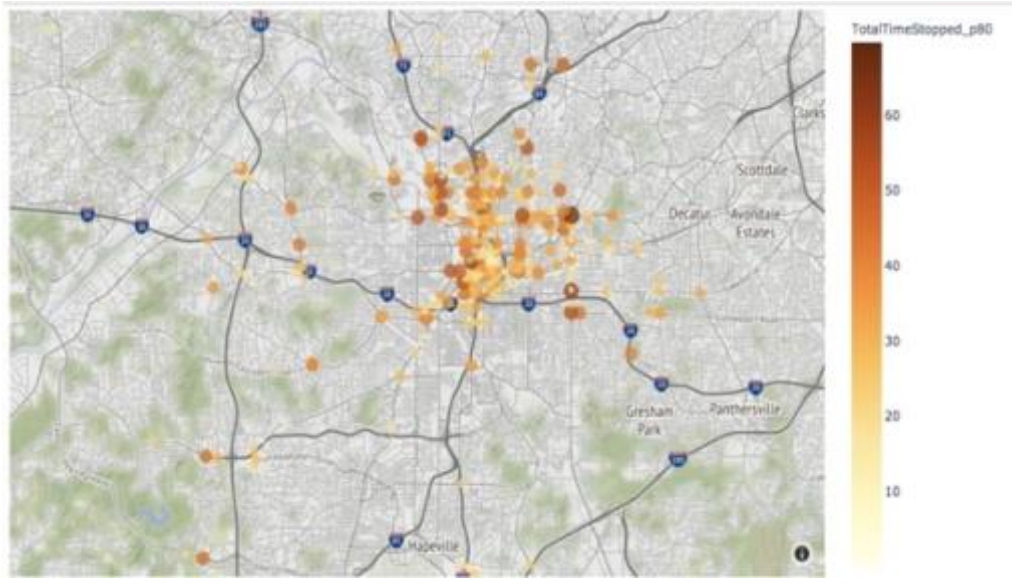

*Figure 3: Busy intersections in Atlanta.*

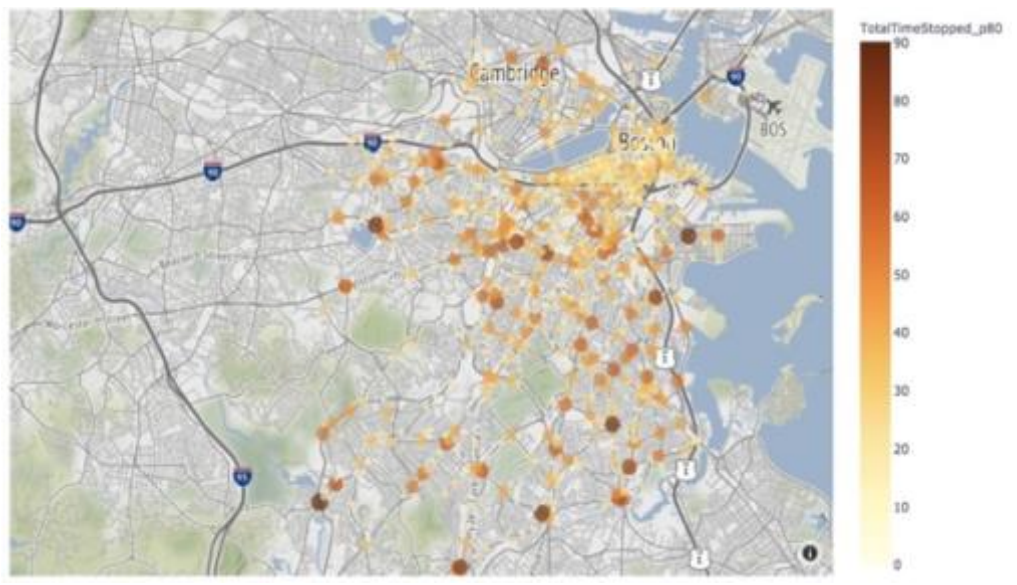

*Figure 4: Busy intersections in Boston.*

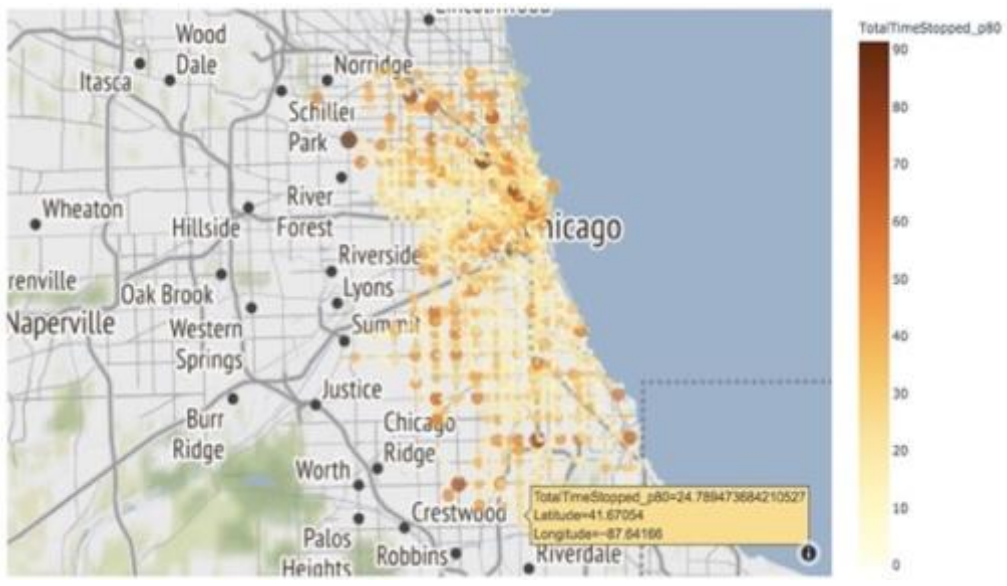

*Figure 5: Busy intersections in Chicago.*

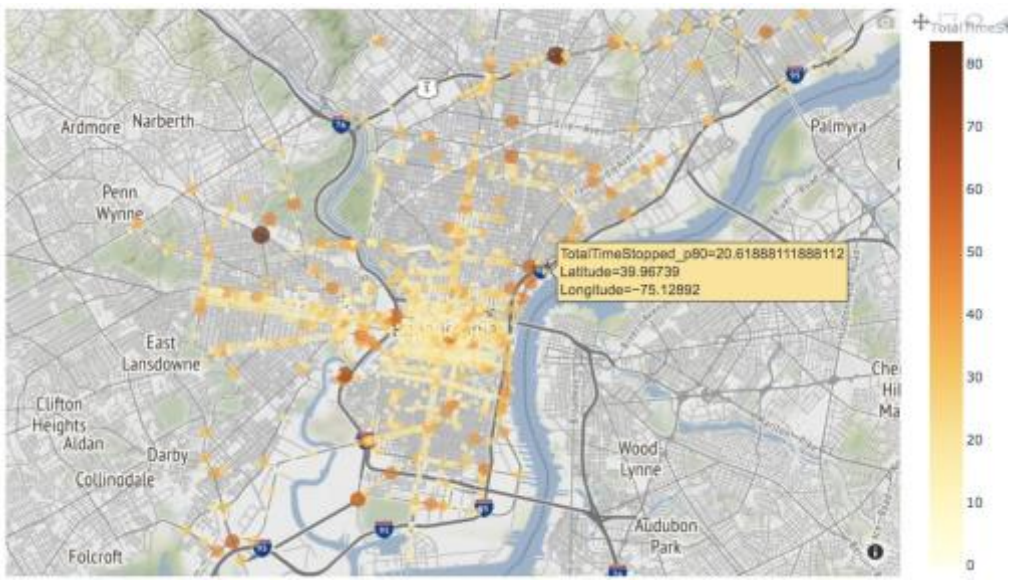

*Figure 6: Busy intersections in Philadelphia.*

# 5.3 Machine Learning Modeling

## Technique 1: Multiple Linear Regression

Multiple linear regression generalizes linear regression, allowing the dependent variable(Y) to be a linear function of multiple independent variables (X's) (Figure 7).

$$Y_i = \beta_0 + \beta_1 X_{1i} + \ldots + \beta_k X_{ki} + \epsilon_i$$

*Figure 7: Multiple linear regression formula.*

The validity of the model depends on whether the assumptions of the linear regression model are satisfied. These are:

- The independent variable is not random. This is true in our case.
- The variance of the error term is constant across observations. This is important for evaluating the goodness of the fit.
- Errors are non-autocorrelated. For this, the Durbin-Watson statistic must be close to 2 which is true in our analysis.

- Errors are normally distributed. If not, then we can’t use some of the statistics, such as the F-test.
- There is no multicollinearity. We ensure this by looking at the correlation matrix of features.

All these assumptions were rigorously tested by us and then the model was further used. To decide on a good model, we used stepwise regression. We tested plenty of combinations of features by adding or removing them one at a time. We added or removed these features based on our logical understanding and by looking at the dataset. Finally, we selected the one that resulted in the best quality which we interpreted using AIC and BIC. In our study, the outliers are anomalies that are important as they are the ones which are actual congestions. We implemented our model by using different loss and regularization functions with different parameters. The Huber loss is a loss function used in robust regression, that is less sensitive to outliers in data. On the other hand, L2 loss is more sensitive to outliers and provides a more stable and closed form solution. We tested first, Huber loss with low values of the parameter delta. This is so that the outliers get penalized according to L1 and other smaller values are penalized according to L2. We further tested L2 loss as well. We finally proceeded with Huber loss with a low value of delta along with little regularization based on our results. Additionally, implementing cross validation enhanced our confidence in the mode (Figure 8).

| | MSE | MAE | Max Error |
|---|---|---|---|
| L2 | 226.8210 | 9.9459 | 297.3281 |
| Huber | 301.2180 | 7.6911 | 309.9721 |

*Figure 8: Loss functions comparison.*

Result and Applications: Linear regression is simple and easy to interpret, and it takes

O(1) constant computation time for prediction. It could be used for various tasks. Huber Loss works well on the average waiting time prediction. The mean absolute error is only around 7.69 minutes, compared to 9.95 minutes from Least Squared Errors. On the other hand, Huber Loss method's errors for the outliers are significantly higher than the Least Squared Errors methods. However, the outliers tend to be rare occasions. Thus, we care more about the average waiting time than the specific outliers in this model. In this case, we are more confident to use Huber Loss to predict the 50-percentile waiting time.

## Technique 2: K-Nearest Neighbors

KNN is a distance-based algorithm. It is a non-parametric algorithm, which means it does not make any assumption on the underlying data distribution. The only assumption it makes is that the data points are in a metric space and thus they have a notion of distance. Typically, we use KNN algorithm for classification problems, but it could be used for regression problems as well. In our problem, it takes the average of a data point's k nearest neighbor's waiting time as the prediction. This is valid because our data lay in a feature space in its nature and using a distance-based algorithm is intuitive. The following photos are the trend of the validation mean squared error, mean absolute error, and max error as K increases for our baseline model. As we could see the trend, the errors start dropping first then start increasing again.

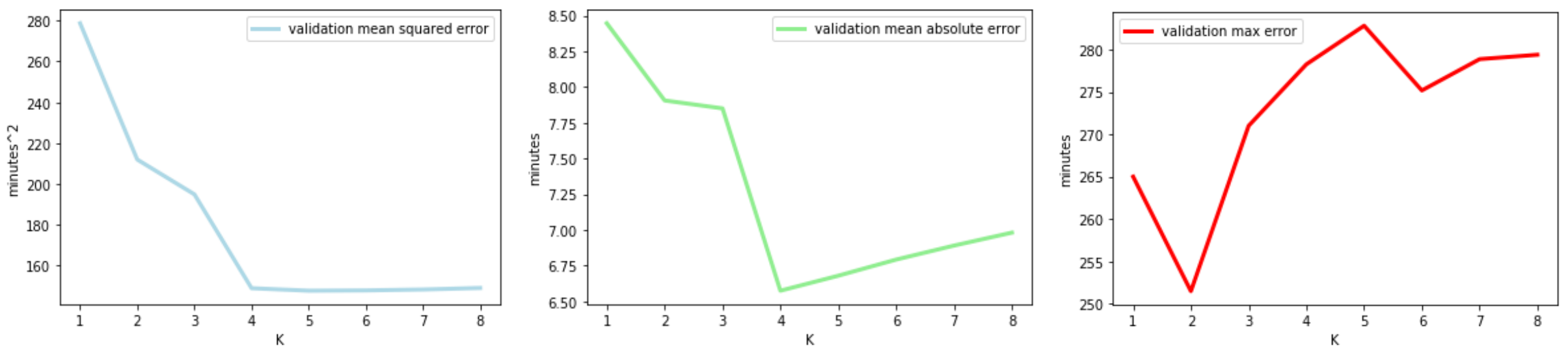

*Figure 9: Validation mean squared error, mean absolute error, and max error.*

We used grid search to find the best combination of our parameters, and the best combination is using uniform weights, and Euclidean metric. After some tuning and feature selection, we find the following two models to have the best validation results (Figure 10).

| | MSE | MAE | Max Error |
|---|---|---|---|
| K = 7 | 135.4573 | 6.1995 | 260.0 |
| K = 8 | 134.8533 | 6.2021 | 263.875 |

*Figure 10: The two models that have the best validation results.*

Therefore, on average we are about 6 mins off the real traffic time at each intersection, but this is largely due to the outliers as we see maximum error.

Application: KNN provides a good prediction on our problem, and it's simple and easy to interpret. One concern of the algorithm is that it takes a long time to run as its computation requires O(n) runtime. This makes the algorithm not practical for real time traffic prediction usage, but it can still serve as a good tool for city traffic analysis and civil engineers planning.

## Technique 3: Gradient Boosting

Gradient boosting is a machine learning technique for regression and classification problems, which makes a prediction model as an ensemble of week prediction models. Boosting is an ensemble technique in which the predictors are made sequentially. The subsequent predictors learn from the mistakes of the previous predictors. The intuition behind gradient boosting algorithm is to repetitively leverage the patterns in residuals and strengthen a model with weak predictions and make it better. The only assumption it makes it's the independence of the observations. This is satisfied as the traffic in one intersection does not depend on another intersection. They could be correlated but our observations on each intersection are independent. For our implementation, we used MSE loss as the other two algorithms.

Results: We tested on different parameters using grid search, and we found the best model with parameters [n_estimators=100, max_depth=3, learning_rate = 0.1, loss='ls']. As we will discuss later in the feature importance, we find that our weather data is not useful for prediction, and we run the model again without weather data (Figure 11).

| | MSE | MAE | Max Error |
|---|---|---|---|
| With weather data | 211.6662 | 9.4953 | 317.9179 |
| Without weather data | 211.5094 | 9.4929 | 325.3719 |

*Figure 11: With or without weather data error comparisons.*

Thus, on average we are about 10 mins off the real traffic time at each intersection, but this is largely due to the outliers as we see maximum error.

Application: Gradient Boosting does not have as good performance as KNN algorithm, but the runtime for prediction is much less. Its prediction computation complexity is only $O(pn_{estimators})$ where p is the number of test data points and $n_{estimators}$ is the number of estimators (we used 100), and it does not depend on the number of training data so the prediction can be done in much less time. So, the Gradient Boosting algorithm could be used for real time prediction to get an approximate on which intersections could be busy at the time, or it can be used as a tool for other algorithms to find the feature importance of each feature. We will discuss our feature importance in the next section.

## 6.0 Discussion

In this study, we explored three different machine learning techniques for predicting traffic congestion at urban intersections: multiple linear regression, k-nearest neighbors (KNN), and gradient boosting. Each approach has its own strengths and limitations, and the choice of technique depends on the specific requirements of the problem and the underlying assumptions about the data.

Multiple linear regression is a widely used technique for modeling the relationship between a dependent variable and one or more independent variables (Hastie et al., 2009). In our case, we used multiple linear regression to predict the average waiting time at intersections based on various features such as the time of day, day of the week, and intersection characteristics. The key advantage of this approach is its simplicity and interpretability. The model assumptions were rigorously tested, and stepwise regression was employed to select the best combination of features based on information criteria (AIC and BIC).

To handle outliers, which represent actual congestion events, we implemented the model using different loss and regularization functions. The Huber loss function, which provides a trade-off between L1 and L2 loss, was found to be effective in reducing the impact of outliers while maintaining good performance for non-outlier instances. Cross-validation further enhanced our confidence in the model's robustness.

The KNN algorithm is a non-parametric, distance-based approach that makes predictions based on the k nearest data points in the feature space (Hastie et al., 2009). In our case, the average waiting time of a data point was predicted using the average of its k nearest neighbors. This method is intuitive and does not make assumptions about the underlying data distribution.

After parameter tuning and feature selection, we identified two KNN models ($k = 7$ and $k = 8$) that performed well on the validation set, with a mean absolute error of approximately 6 minutes. However, the maximum error was relatively high, indicating that the model struggled with some outlier instances. One limitation of the KNN algorithm is its computational complexity, which scales linearly with the number of training instances, making it less practical for real-time traffic prediction.

Gradient boosting is an ensemble technique that combines multiple weak prediction models to create a strong predictive model (Friedman, 2001). It is a powerful algorithm that can handle complex, non-linear relationships in the data and is less susceptible to overfitting compared to other techniques.

In our implementation, we used the mean squared error (MSE) loss function and performed grid search to find the optimal hyperparameters. Interestingly, we found that the inclusion of weather data did not significantly improve the model's performance, suggesting that other factors were more influential in predicting traffic congestion.

The gradient boosting model achieved a mean absolute error of approximately 10 minutes, which is slightly higher than the KNN models. However, its computational complexity for prediction is lower, making it more suitable for real-time applications or as a feature importance tool for other algorithms.

## 7.0 Conclusion

This study presents a comprehensive evaluation of three machine learning techniques for predicting traffic congestion at urban intersections. Each approach offers distinct advantages and trade-offs in terms of performance, interpretability, and computational complexity. The multiple linear regression model, while simple and interpretable, may struggle with capturing complex non-linear relationships in the data. The KNN algorithm demonstrated promising performance on the validation set, with a relatively low mean absolute error. However, its high computational complexity during prediction may limit its applicability for real-time traffic prediction scenarios. Gradient boosting emerged as a powerful technique that can handle complex relationships in the data while maintaining reasonable computational complexity during prediction. Its ability to identify important features also makes it a valuable tool for understanding the underlying factors influencing traffic congestion.

Future research could explore ensemble methods that combine the strengths of these techniques or investigate more advanced deep learning models that can capture spatiotemporal patterns in traffic data. Additionally, incorporating real-time traffic data and integrating the models into intelligent transportation systems could further enhance their practical applicability. However, as the application of machine learning models becomes more widespread, data security emerges as a critical concern. As highlighted in the comprehensive study on cyber security indexes and data protection measures across 193 nations, fortifying the global data fortress requires a multidimensional approach that addresses technical, legal, and policy aspects of data security (Weng et al., 2024). Their findings underscore the importance of implementing robust data protection measures, such as encryption, access controls, and secure data management practices, to safeguard the integrity and confidentiality of sensitive information, including traffic data.

Moreover, the ability to accurately detect and track partially occluded objects is crucial in traffic management systems, as highlighted by the TAO-Amodal benchmark (Hsieh et al., 2023). This benchmark demonstrates the significance of amodal perception, the ability to comprehend complete object structures from partial visibility, in applications like autonomous driving. Techniques like self-training with label-feature consistency that can improve amodal perception by addressing issues with unreliable pseudo-labels could potentially enhance safety in autonomous vehicle deployments (Xin et al., 2023).

Failure to prioritize data security and amodal perception could potentially undermine the benefits of data-driven solutions and compromise public trust. Therefore, future research in this domain should proactively address data security considerations and explore techniques for amodal object detection and tracking, aligning with best practices and adopting a holistic approach to ensure the responsible and secure use of data in traffic management systems. Overall, this study highlights the potential of data-driven approaches in addressing the challenge of urban traffic congestion and provides a foundation for further research and development in this important domain, while emphasizing the need for stringent data security measures and amodal perception capabilities to protect privacy, maintain public trust, and ensure reliable traffic management.